\documentclass[letterpaper, 10 pt, conference]{ieeeconf}  % Comment this line out if you need a4paper

\IEEEoverridecommandlockouts                              % This command is only needed if 
\usepackage[T1]{fontenc}

\usepackage[space, compress, sort]{cite}

\usepackage{amssymb} 
\usepackage{amsfonts}
\usepackage{amsthm}
\usepackage{wrapfig}
\usepackage[ruled,vlined]{algorithm2e}
\usepackage{mathtools}
\usepackage{tabularx}
\usepackage{graphicx}
\usepackage{enumerate}
\usepackage{float}
\usepackage{url}
\usepackage{verbatim}
\usepackage{booktabs} 
\usepackage{multirow}
\usepackage[linkcolor=black,citecolor=goldenrod,urlcolor=black,colorlinks=true]{hyperref}
\usepackage{bm}
\usepackage{cleveref}
\usepackage{subcaption}
\usepackage{caption}      % optional, improves caption formatting
\usepackage{algpseudocode}

\SetKwRepeat{Do}{do}{while}

\usepackage{xcolor}
\usepackage[normalem]{ulem}

\title{\LARGE \bf
SeqVLA: Sequential Task Execution for Long-Horizon Manipulation with Completion-Aware Vision-Language-Action Model
}

\newif\ifanonymous
\anonymoustrue% Set to \anonymousfalse for final version

\author{Ran Yang$^{1\star}$, Zijian An$^{2\star}$, Lifeng Zhou$^{2}$, and Yiming Feng$^{1\dagger}$% <-this % stops a space
\thanks{$^{1}$Ran Yang and Yiming Feng are with Virginia Seafood Agricultural Research and Extension Center, and Department of Biological Systems Engineering, Virginia Tech, 15 Rudd Ln, Hampton, VA, 23669, USA
        {\tt\small \{ryang17,yimingfeng\}@vt.edu}}%
\thanks{$^{2}$Zijian An and Lifeng Zhou are with the Department of Electrical and Computer Engineering,
        Drexel University, 3141 Chestnut St, Philadelphia, PA 19104, USA
        {\tt\small \{za382, lz457\}@drexel.edu}}%
\thanks{$\star$ Equally contributed}%
\thanks{$\dagger$ Corresponding author}%
 }

\begin{document}

\maketitle
\thispagestyle{empty}
\pagestyle{empty}

%%%%%%%%%%%%%%%%%%%%%%%%%%%%%%%%%%%%%%%%%%%%%%%%%%%%%%%%%%%%%%%%%%%%%%%%%%%%%%%%
\begin{abstract}
Long-horizon robotic manipulation tasks require executing multiple interdependent subtasks in strict sequence, where errors in detecting subtask completion can cascade into downstream failures. Existing Vision-Language-Action (VLA) models such as $\pi_0$ excel at continuous low-level control but lack an internal signal for identifying when a subtask has finished, making them brittle in sequential settings. We propose SeqVLA, a completion-aware extension of $\pi_0$ that augments the base architecture with a lightweight detection head perceiving whether the current subtask is complete. This dual-head design enables SeqVLA not only to generate manipulation actions but also to autonomously trigger transitions between subtasks. We investigate four finetuning strategies that vary in how the action and detection heads are optimized (joint vs. sequential finetuning) and how pretrained knowledge is preserved (full finetuning vs. frozen backbone). Experiments are performed on two multi-stage tasks: salad packing with seven distinct subtasks and candy packing with four distinct subtasks. Results show that SeqVLA significantly outperforms the baseline $\pi_0$ and other strong baselines in overall success rate. In particular, joint finetuning with an unfrozen backbone yields the most decisive and statistically reliable completion predictions, eliminating sequence-related failures and enabling robust long-horizon execution. Our results highlight the importance of coupling action generation with subtask-aware detection for scalable sequential manipulation.
\end{abstract}

%%%%%%%%%%%%%%%%%%%%%%%%%%%%%%%%%%%%%%%%%%%%%%%%%%%%%%%%%%%%%%%%%%%%%%%%%%%%%%%%
\section{INTRODUCTION}

Robotic manipulation in complex tasks often involves executing long-horizon tasks that require precise temporal sequencing. Such sequential dependencies arise in many real-world applications: when preparing a meal, a robot must first retrieve a plate, then place utensils, followed by arranging food items in a specific order; in kit assembly, it must insert components into a fixture in the correct sequence, where skipping or misordering steps leads to failure; in laboratory automation, liquid handling tasks must follow an exact pipetting and dispensing order to preserve chemical validity.
These settings exhibit sparse supervision, implicit subgoal transitions, and strong error propagation across stages. A key challenge is enabling manipulation policies to detect subtask completion and determine when and how to transition to the next stage \cite{kroemer2021review, hussein2023detecting}. Naive end-to-end models often struggle under such constraints, lacking a reliable internal signal to segment the task or recover from temporal drift.

Vision-Language-Action (VLA) models are based on large models pretrained on vast robotic datasets and then fine-tuned on domain-specific robotic data to ground instructions in perception and control. They have shown strong capability in mapping rich multimodal context to action sequences with semantic alignment, enabling general-purpose control in a variety of single-stage settings \cite{black2410pi0, driess2023palm, ahn2022can, brohan2022rt,zitkovich2023rt, jiang2022vima, zhao2023learning, zhao2025cot}. These models have been used to support robots in performing practical behaviors such as folding clothes, placing objects, and other structured manipulation tasks. However, their vanilla formulation lacks an internal notion of subtask completion, making them brittle when naively chained for long-horizon, sequential objectives: they either switch prematurely, linger redundantly, or propagate early-stage mistakes into downstream failure or perform in the wrong order \cite{willibald2025hierarchical, huang2023value}.

Classical approaches to long-horizon control often impose structure to handle sequencing. Modern methods handle long horizons using temporal abstraction and hierarchy. Instead of fixed options, they learn hierarchical world models and use goal-conditioned planners to operate at multiple timescales \cite{schiewer2024exploring, wang2023goplan}. Related methods aim to determine when a subtask or goal is completed. Learned termination critics, progress monitors, and vision-language success detectors have been used to give adaptive stop signals or catch failures in multi-stage policies \cite{duan2024aha, gu2025safe}. Despite these advances, termination signals are usually kept separate from action generation, and hierarchical or planning systems still depend on external controllers or hand-specified intermediate goals to sequence stages. As a result, a gap remains: how to endow a VLA generation model with a tightly coupled, learned signal that detects when each one is complete, allowing it to autonomously trigger the next stage and self-pace the chaining of dependent subtasks without external control. 

% \lz{LZ: here several prior studies are mentioned, we will need to either compare with them or justify why we don't}
% \lz{LZ: I think we first use an LLM to decompose a task into a sequence of subtasks and then implement pi0 with completion. We need to discuss if we want to add this module.}

In this paper, we bridge this gap with Sequential-VLA (SeqVLA), which is built upon the $\pi_0$ framework and features a \textbf{Completion Detection Head}. Given the same visual, linguistic, and action-history context, this head predicts whether the current subtask has been successfully achieved. We explore finetuning strategies by freezing different sections of $\pi_0$ to identify which components need adaptation for reliable sequential execution. We then evaluate the proposed model on two distinct scenarios, demonstrating substantial improvements in overall success rate compared to the original $\pi_0$ and several strong baselines.
Our contributions are:
\begin{itemize}
\item We integrate a learned task completion detection head into the $\pi_0$ model, enabling the model to infer subtask completion from multi-modal context.
\item We identify the most effective finetuning strategy for the augmented VLA by freezing different sections of $\pi_0$, demonstrating how to adapt the model to generate correct subtasks in sequence reliably.
\item We evaluate the resulting framework in two real-world sequential scenarios and compare it with the strong baseline, demonstrating that our approach yields substantially superior task-level performance.
\end{itemize}
% real-world manipulation task such as.. often involves sequential long horizon... 

% the start-of-art vla .... pi0 long-horizon 
% however, order may not be correct 

% we propose ...

% results.... 

\section{Related Work}
\subsection{Vision-Language-Action Models}

Vision-Language-Action (VLA) models aim to unify visual perception, language understanding, and continuous robot control within a single generative framework. Instead of decoupling language grounding, visual recognition, and action synthesis into separate modules, VLA systems directly learn the conditional distribution.
\[
p(\mathbf{a}_{1:T} \mid \mathbf{o}_{1:T}, \mathbf{l}),
\]
over action sequences $\mathbf{a}_{1:T}$, conditioned on image and state observations $\mathbf{o}_{1:T}$ and a natural language prompt $\mathbf{l}$. This joint modeling approach enables better semantic alignment between high-level instructions and low-level behaviors, particularly in open-world or zero-shot settings.

Most VLA systems employ a vision-language encoder backbone, such as BLIP \cite{li2022blip} or SigLIP \cite{zhai2023sigmoid}, to embed both the language command and visual observations into a shared representation space. The output of the VLM is then passed to an action-generation module, which produces either continuous control commands or discrete action tokens depending on the model architecture.

The core distinction between different VLAs lies in how they generate action sequences. One class of models, exemplified by SayCan \cite{ahn2022can} and PaLM-E \cite{driess2023palm}, decomposes tasks into high-level symbolic skills and uses large language models to plan over these skills before grounding them into low-level controllers. This symbolic planning approach facilitates semantic reasoning, but often relies on predefined skill libraries and structured environments.
A second category of methods, such as RT-1 \cite{brohan2022rt}, RT-2 \cite{zitkovich2023rt}, and OpenVLA \cite{kim2024openvla}, formulates action prediction as a token-by-token autoregressive generation task. These models discretize continuous actions into token sequences and train transformers to decode them from vision-language context. 
A third class of approaches, including VIMA \cite{jiang2022vima}, CoT-VLA \cite{zhao2025cot,zhao2023learning}, and SmoVLA \cite{shukor2025smolvla}, leverages latent trajectory modeling, contrastive learning, or smooth policy regularization in the latent space. These models typically embed demonstrations into compact latent codes and generate behavior via infilling or trajectory decoding. 
% Despite architectural differences, most VLAs struggle with long-horizon, sequential tasks in which the agent must implicitly detect subgoal completion and determine when to transition between stages. These challenges motivate our investigation into extending VLA models with subtask-aware control mechanisms.

\subsection{The \texorpdfstring{$\pi_0$}{pi0} Model}

The $\pi_0$ model \cite{black2410pi0} is a representative VLA framework that emphasizes continuous high-frequency robotic control, with a particular focus on robustness to horizon length and ambiguity. Unlike models that decode action sequences token-by-token or operate in a latent trajectory space, $\pi_0$ directly learns a continuous function $\pi_\theta(\mathbf{a}_t \mid \mathbf{z}_t)$ to predict actions based on a shared observation-language context $\mathbf{z}_t$.

The core of $\pi_0$ is a flow matching strategy for trajectory supervision. During finetuning, the model first samples a noisy initial action $\mathbf{a}_t^{(0)}$ and a ground-truth action $\mathbf{a}_t^{(1)}$ from human demonstration. A random scalar $\tau \in [0,1]$ is drawn to interpolate between them via a straight-line path:
\begin{equation*}
    \mathbf{a}_t^{(\tau)} = (1 - \tau) \cdot \mathbf{a}_t^{(0)} + \tau \cdot \mathbf{a}_t^{(1)}.
\end{equation*}
The model is trained to predict the velocity vector pointing from $\mathbf{a}_t^{(0)}$ to $\mathbf{a}_t^{(1)}$ using the intermediate $\mathbf{a}_t^{(\tau)}$ and conditioning context, resulting in the following supervised loss:
\begin{equation*}
    \mathcal{L}_{\text{flow}} = \mathbb{E}_{\tau, \mathbf{a}_t^{(0)}, \mathbf{a}_t^{(1)}}\left[\left\| \pi_\theta(\mathbf{a}_t^{(\tau)}, \tau, \mathbf{z}_t) - (\mathbf{a}_t^{(1)} - \mathbf{a}_t^{(0)}) \right\|_2^2\right].
\end{equation*}
This formulation allows the policy to match the target velocity field over the entire trajectory manifold rather than only at the final endpoints, offering smoother control and better resilience to noise or partial observability.

Compared to token-based autoregressive models, $\pi_0$ benefits from continuous-time reasoning and variable rollout lengths. It does not require action discretization or fixed-length decoding, which helps mitigate error accumulation and temporal drift across long sequences. The flow-based formulation also supports iterative refinement: even if the intermediate predictions are ambiguous, the model can gradually converge toward the goal through continuous updates. These design choices make $\pi_0$ particularly suitable for robotic manipulation tasks that demand both temporal smoothness and precise low-level control, providing a compelling advantage over symbolic or token-based VLAs that often struggle with continuity, compounding errors, or rigid horizon constraints.\footnote{At the time our experiments were conducted, the latest variant $\pi_{0.5}$ \cite{intelligence2025pi05visionlanguageactionmodelopenworld} had been announced but was not yet publicly released. Accordingly, we base our implementation on the most recent available model, $\pi_0$.}

However, despite these advantages, $\pi_0$ operates uniformly over entire action segments through flow matching on observation-language context and lacks an explicit mechanism for identifying or sequencing subtasks. Specifically, the action finetuning is solely based on $\mathbf{z}_t$ without awareness of task completion status during sequential execution. This limitation poses challenges in structured long-horizon manipulation, where the agent must not only produce smooth low-level motion but also recognize when a subtask is complete and transition to the next. To address this, we extend $\pi_0$ with a dedicated completion detection head that monitors subtask status, and a chaining mechanism that enables self-paced progression across sequential stages.

\section{Methods}
\begin{figure*}[thb]
    % \vspace{-3mm}
    \vspace{2mm}
    \centering
    \includegraphics[width=0.96\textwidth]{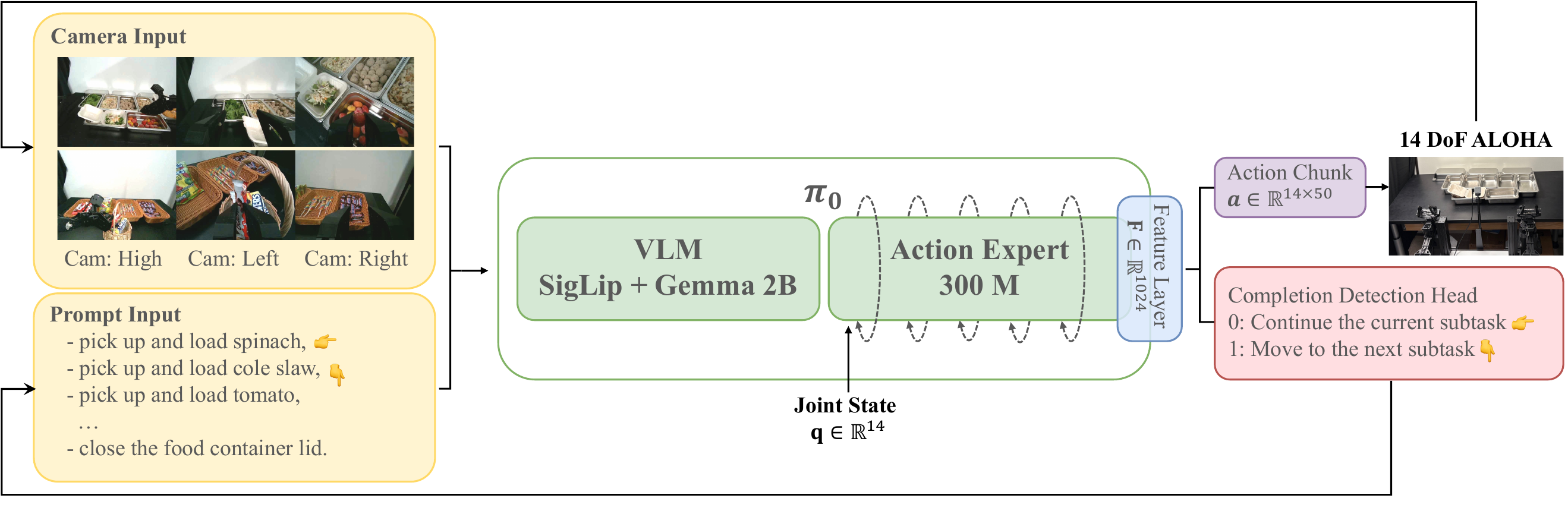}
    \caption{Architecture of SeqVLA. The inputs are the same as $\pi_0$. The completion detection head takes the feature layer from the Action Expert as input to generate the task completion probability.}
    \label{fig:seqVLA_architecture}
\end{figure*}
\subsection{Problem Formulation}
Real-world manipulation tasks, such as food handling or assembly operations, usually require sequential execution of multiple subtasks. 
% For instance, serving a meal from customized order involves systematrically loading different food stuffs to a container, where each subtask needs to be completed before proceeding to the next one.
We fomulate such long-horizon tasks as a sequential order  $ \mathcal{T} = \{\mathcal{T}_1, \mathcal{T}_2,..., \mathcal{T}_n\}$ where the completion of task $\mathcal{T}_i$ is the prerequisite of the initialization of task $\mathcal{T}_{i+1}$. Challenges arise when using existing VLA models, as they may struggle to execute subtasks in the correct order, especially when visual scenes appear similar across different subtask stages. For example, a partially filled food container would resemble a nearly filled one before or after loading the last food type. SeqVLA addresses the task sequence confusion issue by learning not only the action to execute but also when each subtask is completed, which enables the proper execution of long-horizon tasks that consist of multiple subtasks.

\subsection{VLA Architecture with Completion Detection Head}
SeqVLA extends a standard VLA architecture with a lightweight completion detection head, as illustrated in Figure~\ref{fig:seqVLA_architecture}. The model takes multi-view visual observations from three cameras, the current robot joint states, and a language prompt as inputs, and jointly produces two outputs: continuous robot control actions for the current subtask and a binary classification score indicating whether the subtask has been completed. The completion signal then triggers a transition to the next subtask in the sequence, enabling structured long-horizon execution. We implement SeqVLA on top of $\pi_0$~\cite{black2410pi0}, which provides a strong foundation through its SigLIP vision encoder, Gemma-2B language backbone, and Gemma-300M action expert. By leveraging $\pi_0$’s extensive pretraining across diverse robotic tasks and systems, SeqVLA inherits robust low-level control and mistake recovery capabilities. At the same time, our additional completion detection mechanism equips it with the ability to reason over subtask boundaries and enforce correct sequencing.
% \lz{LZ: where is the head, how does it connect to the pi0 module? Draw that in the figure, as this is the main novelty of our work}
SeqVLA extends the base $\pi_0$ model with a parallel completion detection output that shares the exact feature representations from the action expert layer. The original action head retains the function of predicting a 14-DoF output for manipulating a dual-arm Aloha~\cite{fu2024mobile}. Our add-on task-tracking head performs binary classification to determine the completion status of the ongoing subtask. This feature-sharing architecture guarantees computational efficiency during inference while leveraging the information extracted by the powerful pre-trained backbone model. 

Completion detection head, shown in Figure \ref{fig:seqVLA_architecture}, is a binary classifier that transforms the feature representations from the action expert into a probability, which we denote as task state indicator $p$,
\begin{equation*}
    p = \sigma (\textbf{W} \cdot \textbf{F} + b),
    % \label{task classifier}
\end{equation*}
where $\textbf{W}\in \mathbb{R}^{1024}, b\in\mathbb{R}$ are learnable parameters, $\sigma$ is sigmoid activation function $\sigma(x)=1/(1+e^{-x})$, and $\mathbf{F}\in\mathbb{R}^{1024}$ is the generated feature layer from the action expert. We measure the loss of task completion term by binary cross-entropy loss \cite{mao2023cross}:
\begin{equation*}
    L_{\text{completion}}(y,\hat{y})=-[y\cdot log(\hat{y})+(1-y)\cdot log(1-\hat{y})].
\end{equation*}
Accordingly, the loss function is given by $L_\text{total} = L_\text{action} + \lambda \cdot L_\text{completion}$ where $L_\text{action}$ is the standard $\pi_o$ action prediction loss from the 3M action expert. We set $\lambda = 0.1$ as the weight for the completion loss contribution.

% The binary classifier uses a single linear layer that transforms the 1024-dimensional feature representations from the action expert into a completion probability: $c = \sigma(W \cdot f_{action} + b)$ where $f_{\text{action}} \in \mathbb{R}^{1024}$ represents the feature representations from the action expert, $W \in \mathbb{R}^1 \times 1024$ \lz{LZ:?} is the learned weight vector, $b \in R$ is the bias, $\sigma$ is the sigmoid activation function, and $p \in [0,1]$ is the task completion probability. The loss is given by $L_\text{total} = L_\text{action} + \lambda \cdot L_\text{completion}$ where $L_\text{completion} = BCE (pred,gt)$ \lz{LZ:?}, $L_\text{action}$ is the standard $\pi_o$ action prediction loss from the 300M action expert, $L_\text{completion}$ is binary cross-entropy loss for completion classification, and $\lambda = 0.1$ weights the completion loss contribution.

\subsection{Subtask Execution}
SeqVLA offers a simple yet effective solution that leverages the binary head output to manage subtask transitions and maintain the accuracy of the task sequence. It operates by processing each inference to obtain both an action chunk for execution and a completion assessment for transition decisions.
At each inference step, SeqVLA produces two outputs: an action chunk and an execution indicator $p \in [0,1]$, which represents the probability that the agent should continue the current subtask ($p$ close to 1) or terminate it and transition to the next ($p$ close to 0).
 % The controller employs this probability value to determine whether to execute the predicted action chunk or transition to the next subtask. 
 When a subtask completion signal is detected ($p < \tau$), the indicator would initialize a transition process: 1) stop execution of the robot action immediately, 2) send the robot to home pose for consistent task initialization, and 3) begin the following subtask by switching the task prompt. A threshold $\tau = 0.2$ is selected to trigger task transitions based on our empirical results.\footnote{In experiments, it is observed that when the threshold is set too low (e.g., 0.1), the policy becomes overly conservative and rarely issues a stop signal, making it difficult to terminate at the correct time. Conversely, when the threshold is set too high, the policy becomes overly sensitive to minor variations in the scale readings, which often leads to premature stopping and fluctuating behavior.} 
 % This relatively low value provides a safety margin: it avoids premature switching due to noisy confidence estimates near the decision boundary, while ensuring that once the model’s confidence in continuation drops substantially, the transition is triggered promptly to maintain reliable subtask sequencing and safe robot operation
% \lz{LZ:why?}

\section{Experiments}
\subsection{Experimental Setup}
\begin{figure}[!thb]
        \vspace{2mm}
        \centering
        \includegraphics[width=0.4\textwidth]{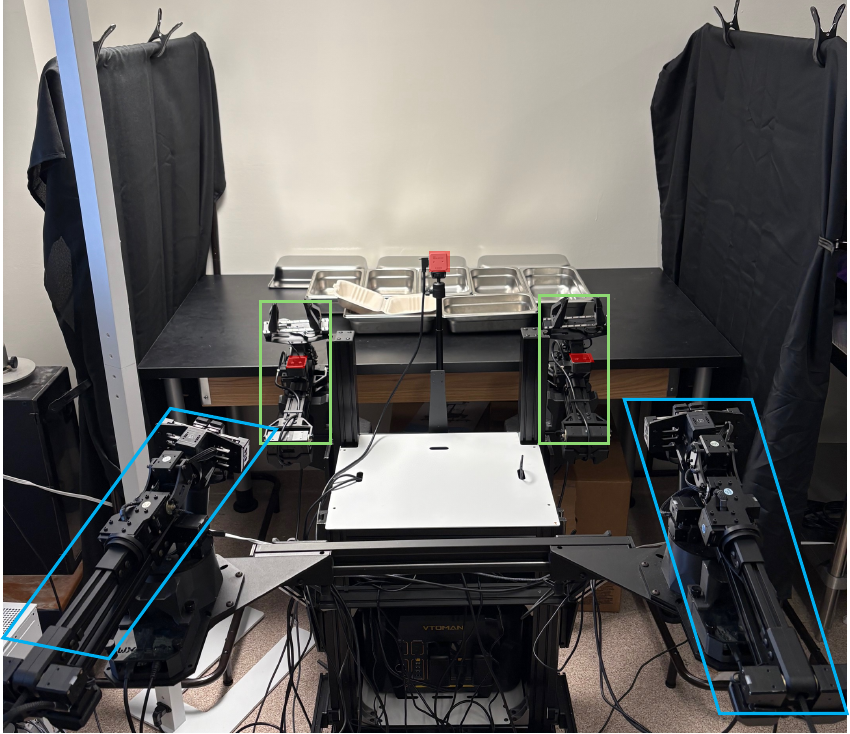}
        \caption{Experiment setup. The three red boxes denote the cameras: two are mounted on the gripper to monitor the conditions at the gripper’s end, and one is positioned in front of the robot to provide a global view. The two green boxes indicate the robotic arms responsible for grasping objects. The two blue boxes correspond to the kinesthetic teaching arms, which are used to guide the motion of the front robotic arms during data collection for finetuning.}
        \label{fig:setup}
\end{figure}

\begin{figure}[!h]
        \centering
        \begin{subfigure}{0.5\textwidth}
        \centering
        \includegraphics[width=\textwidth,trim=0mm 0mm 0mm 1cm, clip]{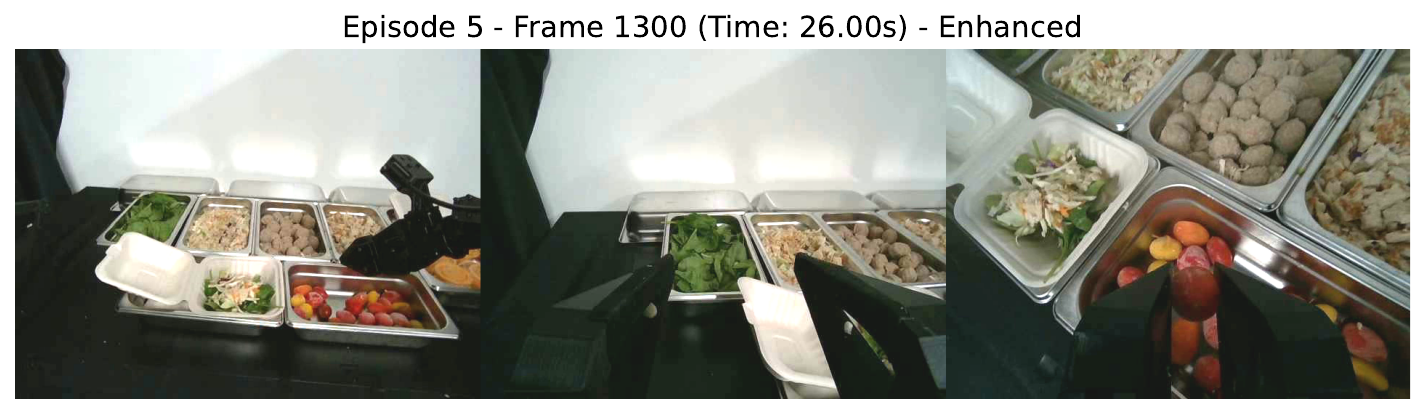}
        \caption{Camera views in salad task}
        \label{fig:cam1}
        \end{subfigure}% 

        \begin{subfigure}{0.5\textwidth}
        \centering
        \includegraphics[width=\textwidth,trim=0mm 0mm 0mm 1cm, clip]{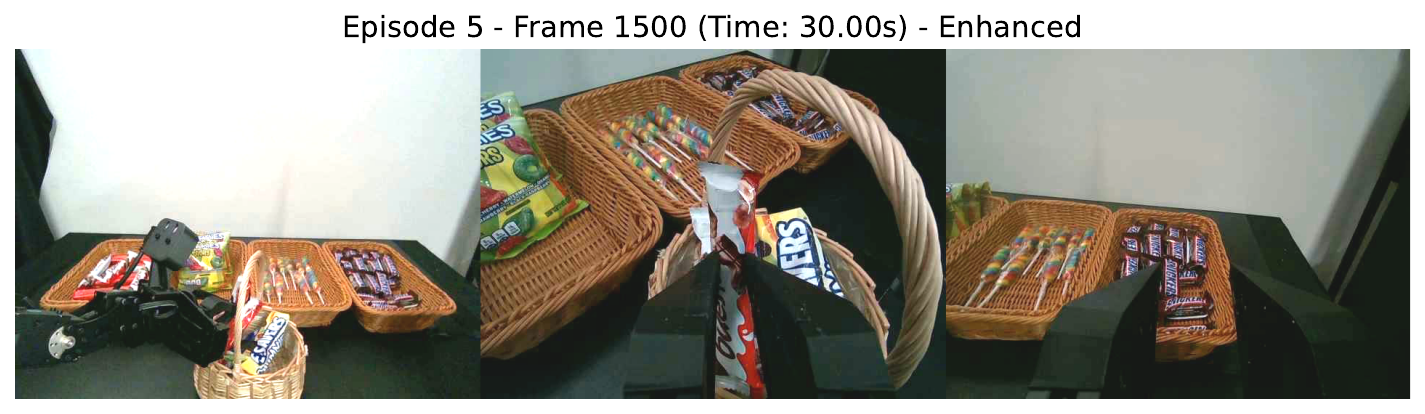}
        \caption{Camera views in candy task.}
        \label{fig:cam2}
        \end{subfigure}%
        \vspace{-1mm}
        \caption{Camera view on two tasks, with high camera (left), left camera (middle), and right camera (right).}
        \label{fig:cam_view}
\end{figure}

\begin{figure}[thb]
        \centering
        \begin{subfigure}{0.48\textwidth}
                \centering
                \includegraphics[width=\textwidth]{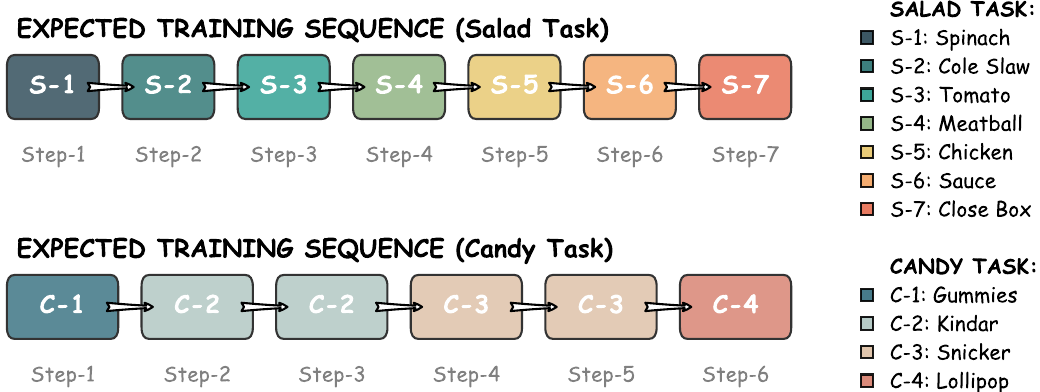}
                \caption{Subtask sequences for long-horizon salad and candy packing tasks.}
                \label{fig:task_sequence_diagram}
        \end{subfigure}%
        \vspace{5mm}

        \begin{subfigure}{0.23\textwidth}
                \centering
                \includegraphics[width=\textwidth]{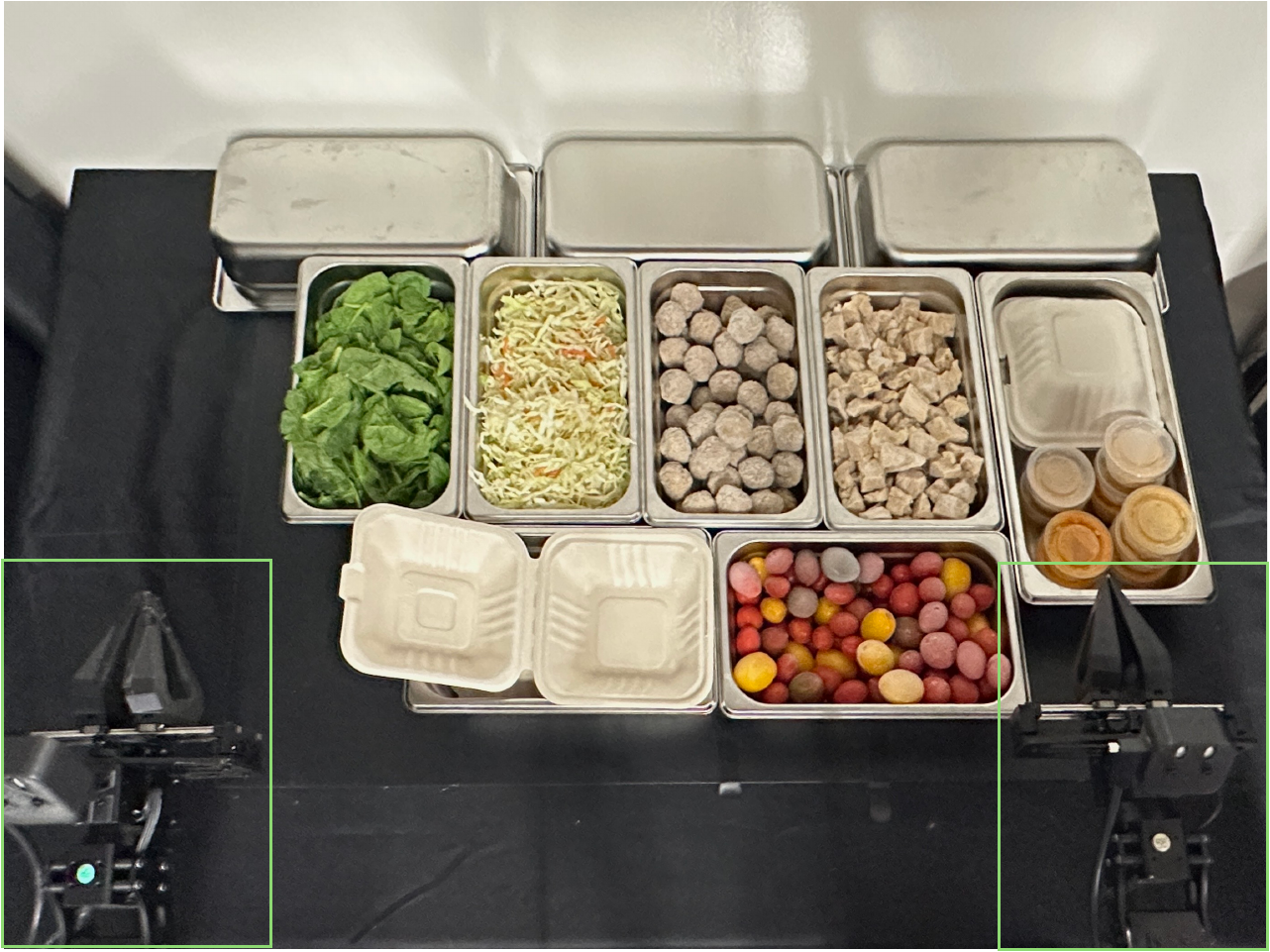}
                \caption{Task1: Salad packing}
                \label{fig:salad_task}
        \end{subfigure}%
        \hfill
        \begin{subfigure}{0.23\textwidth}
                \centering
                \includegraphics[width=\textwidth]{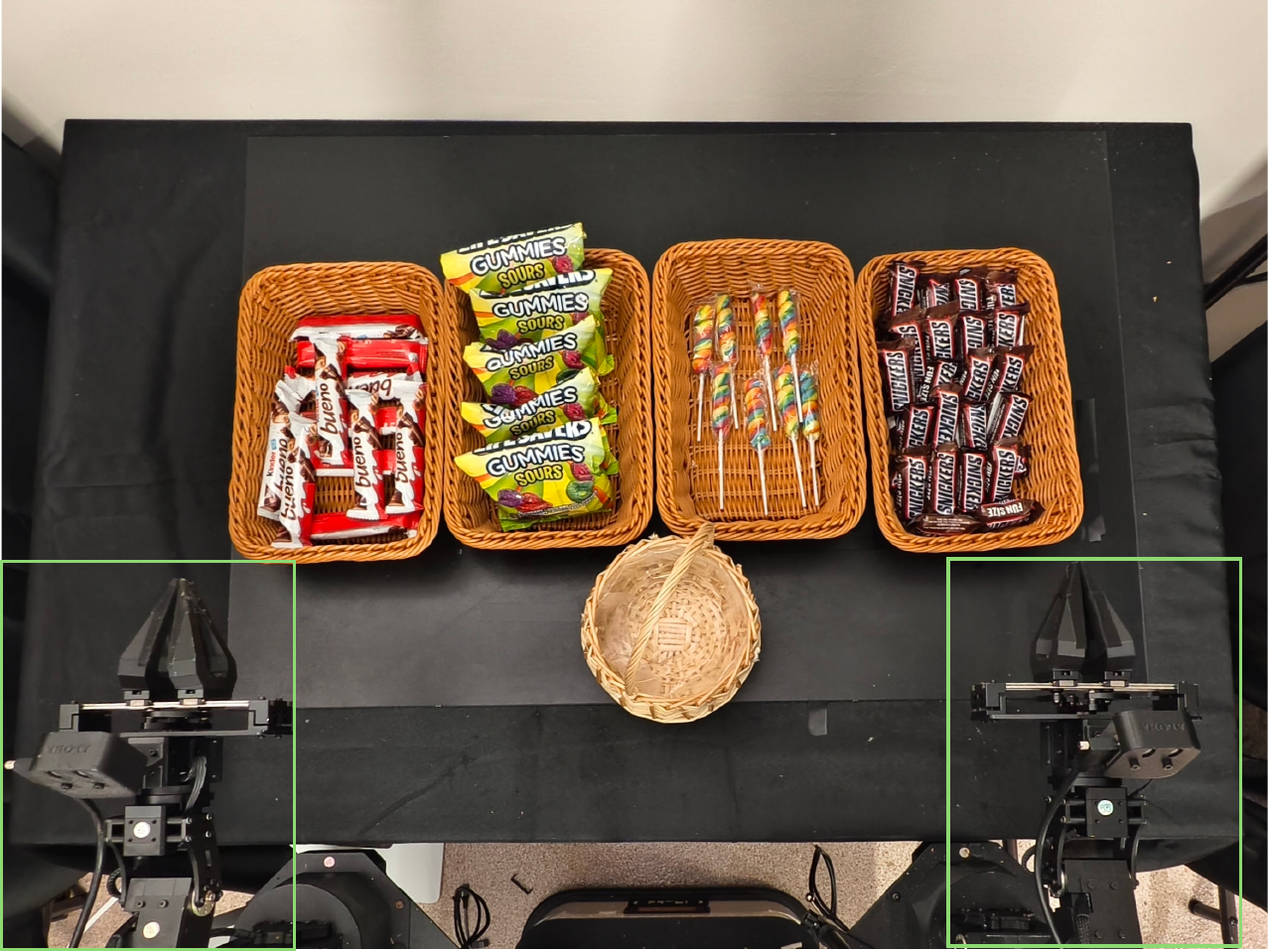}
                \caption{Task2: Candy packing}
                \label{fig:candy_task}
        \end{subfigure}
        \caption{Task setup. (a) expected long-horizon task sequences for the Salad and Candy tasks, showing the sequential order of all subtasks. (b),(c) visualization of the experimental setup for the two tasks, where the green boxes mark the robotic arm.}
        \label{fig:task_sequence}
\end{figure}

Our experiments feature a bimanual 14-DoF Aloha robot, equipped with three cameras for collecting RGB images from central top, left wrist, and right wrist views during teleoperation, as shown in \Cref{fig:setup} and \Cref{fig:cam_view}. We collected two types of teleoperation demonstrations: individual subtask demonstrations for finetuning SeqVLA, and complete long-horizon sequential demonstrations for finetuning the baseline $\pi_0$. Such a data collection strategy enables the precise labeling of subtask completion status for SeqVLA, while also allowing $\pi_0$ to perform the same long-horizon tasks for fair comparison. We present experiments on two different long-horizon tasks, each consisting of multiple subtasks, as illustrated in \Cref{fig:task_sequence_diagram}. The detailed setup for the two experiments is illustrated in \Cref{fig:salad_task} and \Cref{fig:candy_task}. We have attached a video of our experiments to the supplementary material.

\textbf{Salad Packing}: this environment is set up to mimic a food service scenario, where the robot is assigned to load the foodstuffs in the designated order (spinach, coleslaw, meatball, chicken, tomato, and sauce cup) into the food container and then close it. The left arm handles loading tasks for spinach and coleslaw, while the right arm picks up and loads the other items. Container closing requires bimanual coordination for lid placement and securing. This task assesses the ability to maintain the correct sequential order across diverse subtasks involving different objects and manipulations.

\textbf{Candy Packing}: this task examines scenarios where certain subtasks are executed multiple times within a single long-horizon sequence. The robot picks gummies (left arm), places Kinder chocolates twice consecutively (left arm), adds Snickers bars twice (right arm), and finishes with lollipop placement (right arm). This task examines the capability to handle repeated subtask execution within a long-horizon sequence. 


\subsection{Dataset construction}
We collect two complementary types of demonstration data. The first type is subtask-level demonstrations, where each subtask in \Cref{fig:task_sequence_diagram} is paired with a unique language prompt. For example, $S\text{-}1$ is prompted as ``Pick up the spinach,'' while $C\text{-}2$ is prompted as ``Pick up the kinder.'' These demonstrations are used to fine-tune SeqVLA to ground prompts into subtask-specific behaviors and to learn subtask completion detection. The second type is long-horizon demonstrations, where entire salad-packing and candy-packing tasks are executed following the designed subtask sequences. These complete trajectories are used to finetune the baseline VLA $\pi_0$ for end-to-end long-horizon execution.

All subtask-level demonstrations are manually annotated with binary labels to indicate whether each frame still contributes to the execution of the subtask. Frames labeled as 1 correspond to movements that are actively advancing the subtask. In contrast, frames labeled as 0 correspond to non-contributory actions, such as holding the arm stationary after an object has already been placed in the container. Consequently, the label sequence follows patterns such as 1111$\cdots$1100$\cdots$00), where 1s indicate the execution phase and 0s indicate the post-completion phase. In practice, the transition to label 1 is determined by strict criteria: the target object must have been successfully placed in the container, the gripper must be fully released, and there should be no contact between the gripper and the container. These conditions ensure that subsequent reset motions, such as returning the arm to the home pose, do not interfere with the current task state.

\subsection{Finetuning Strategy}
\begin{figure}[h]
        % \vspace{-3mm}
        \centering
        \begin{subfigure}{0.22\textwidth}
        \centering
        \includegraphics[width=\textwidth]{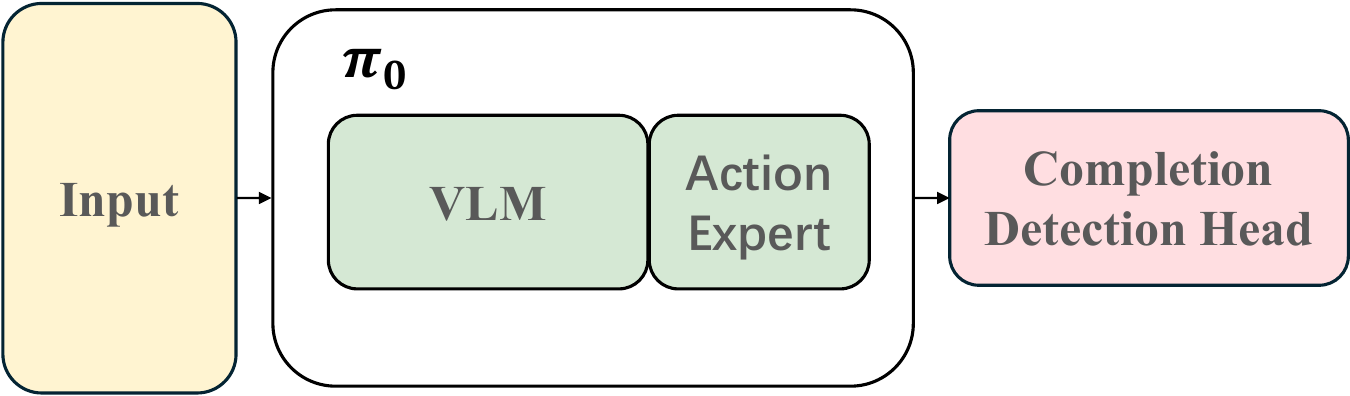}
        \caption{SeqVLA-J}
        \label{fig:subfig1}
        \end{subfigure}% 
        \vspace{3mm}
        \hfill
        \begin{subfigure}{0.22\textwidth}
        \centering
        \includegraphics[width=\textwidth]{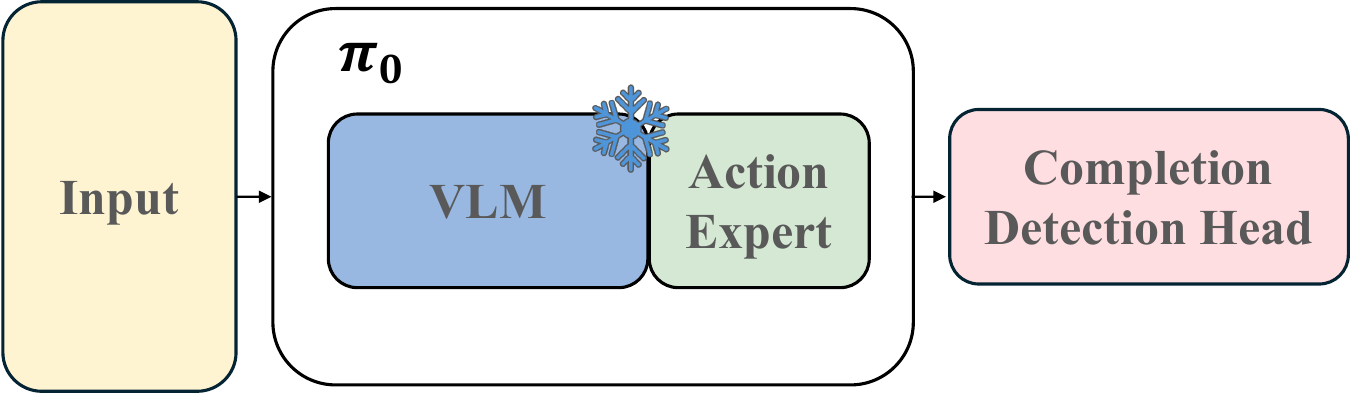}
        \caption{SeqVLA-JF}
        \label{fig:subfig2}
        \end{subfigure}%
        \hfill
        \begin{subfigure}{0.22\textwidth}
        \centering
        \includegraphics[width=\textwidth]{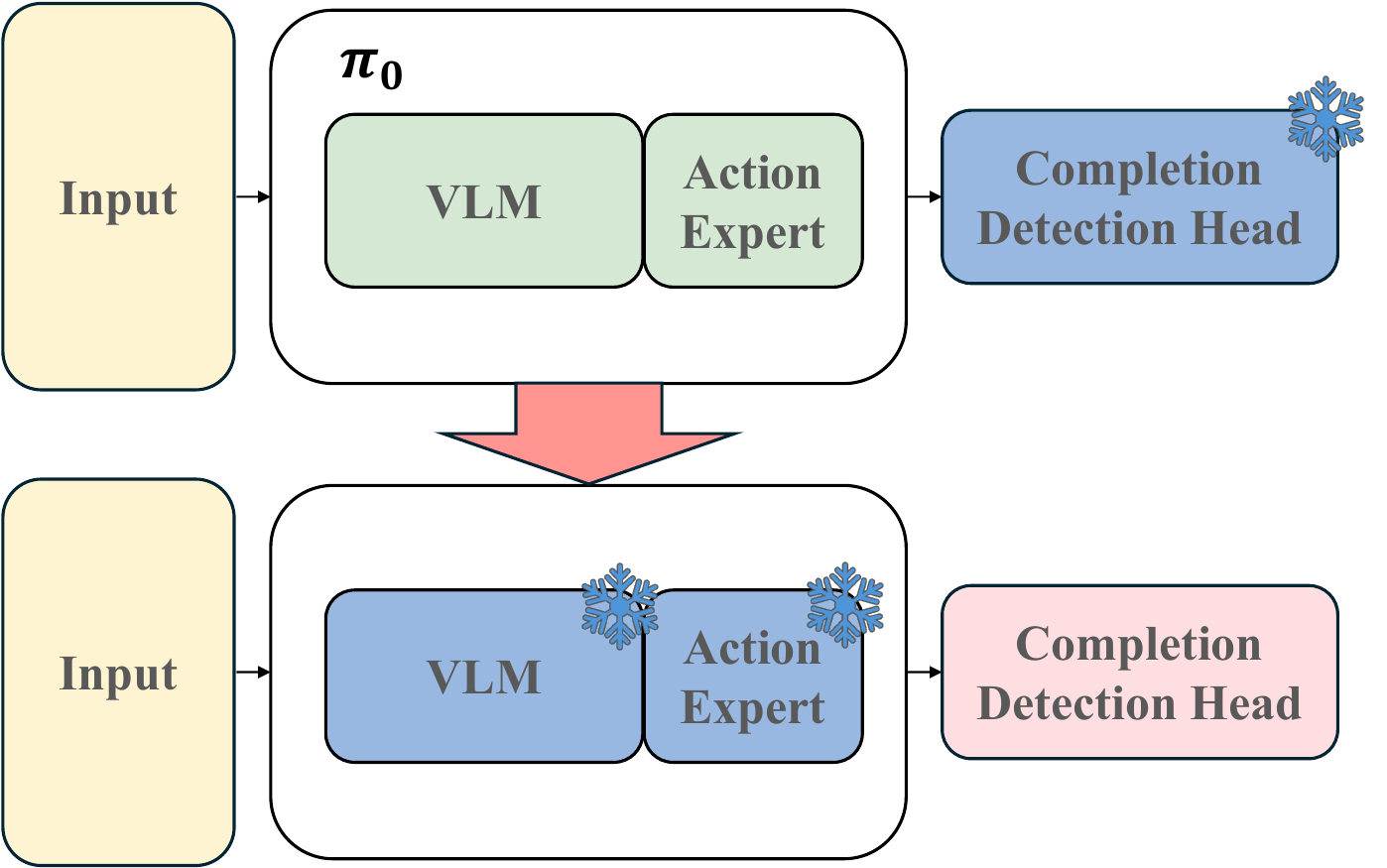}
        \caption{SeqVLA-S}
        \label{fig:subfig3}
        \end{subfigure}%
        \hfill
        \begin{subfigure}{0.22\textwidth}
        \centering
        \includegraphics[width=\textwidth]{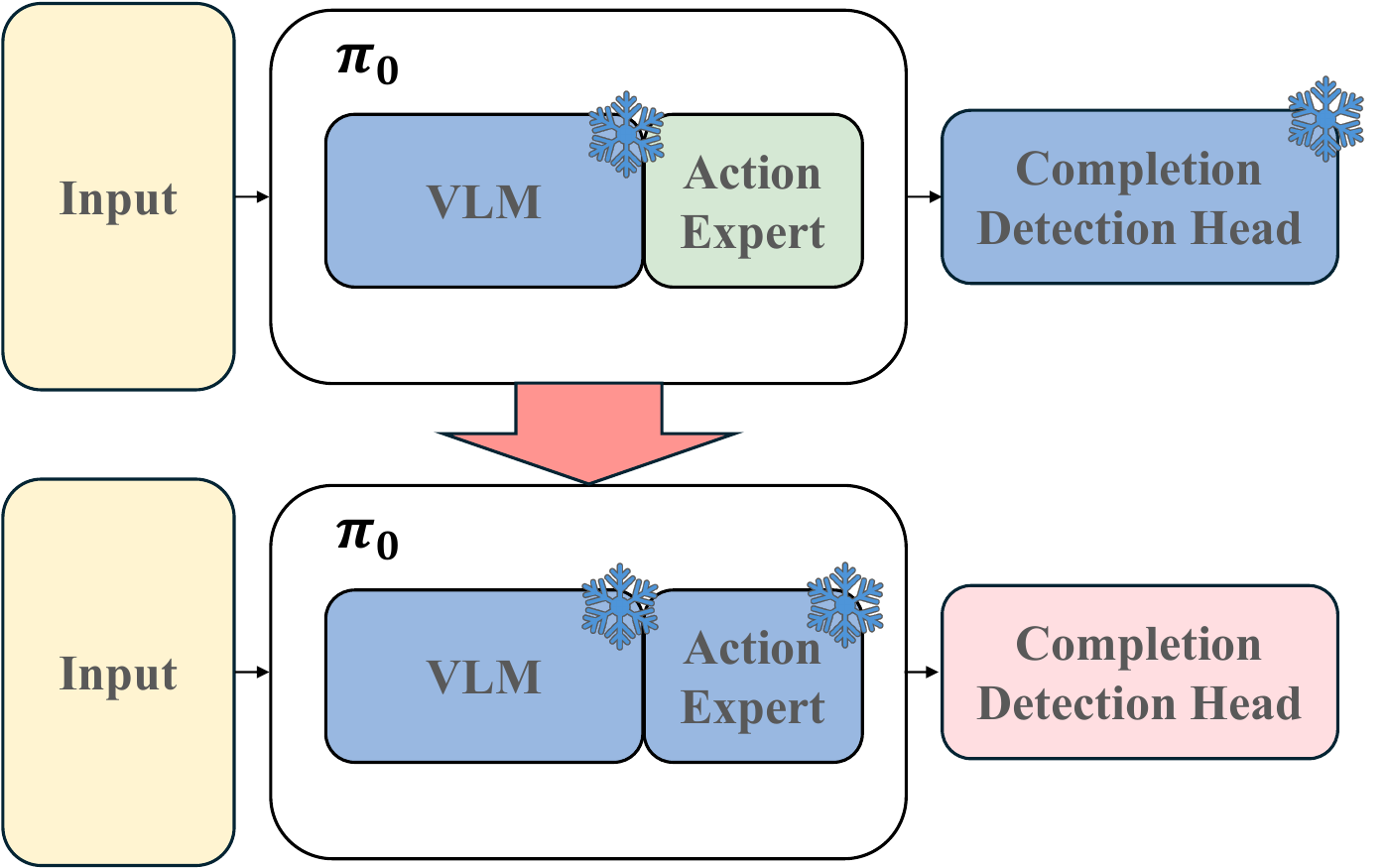}
        \caption{SeqVLA-SF}
        \label{fig:subfig4}
        \end{subfigure}
        
        \caption{Four finetuning strategies.}
        \label{fig:training_strategy}
\end{figure}

Integrating a completion detection head into a pre-trained VLA model raises the question of how to fine-tune the dual-head architecture best so that the model can adapt to predicting both actions and task completion without erasing its pre-trained competencies. To integrate the additional completion-detection head into the pre-trained $\pi_0$ backbone, we consider two orthogonal design dimensions in the finetuning procedure as illustrated in \Cref{fig:training_strategy}. The first dimension concerns how to optimize the dual outputs, balancing action prediction with task-completion classification. The second dimension addresses how to preserve the generalizable knowledge obtained through $\pi_0$’s pretraining while enabling adaptation to our tasks. By combining choices along these two dimensions, we arrive at four distinct finetuning configurations for SeqVLA, which we evaluate systematically.

We first examine different approaches for finetuning the dual outputs of SeqVLA. Since the model must simultaneously predict low-level manipulation actions and high-level task completion signals, a natural option is to fine-tune both heads jointly, allowing the two objectives to inform each other during optimization. However, joint optimization may also introduce conflicts, especially if the classification head is immature in early finetuning and destabilizes action learning. To address this, we also consider a sequential strategy in which the action head and backbone are trained first to ensure stable control, and the classification head is introduced afterward to learn on top of a well-initialized representation. The following outlines these two types of finetuning in more detail.
\begin{itemize}
\item Joint finetuning: Action and classification heads are trained simultaneously for coupled learning of manipulation and task completion detection.
\item Sequential finetuning: Action head is trained first together with the entire backbone, with the classification head frozen; then the entire backbone with the action head is frozen to fine-tune the completion classification head.
\end{itemize}

In addition, we explore different strategies for preserving the knowledge obtained through $\pi_0$’s pretraining while adapting to our tasks. Full fine-tuning allows all parameters, including the pre-trained vision-language backbone, to update, thereby providing maximum flexibility for domain-specific adaptation. However, such aggressive updating risks overwriting the generalizable representations that $\pi_0$ has acquired from large-scale pretraining. As a contrasting strategy, we also consider freezing the pre-trained backbone so that only the action expert and the two prediction heads are updated. This approach helps safeguard the original vision-language grounding ability of $\pi_0$, but may limit task-specific adaptation capacity. 
\begin{itemize}
\item Full finetuning strategy: All parameters, including the pre-trained VLM backbone, are allowed to update during finetuning for domain-specific learning on our tasks.
\item Frozen backbone strategy: Pre-trained VLM backbone is fixed during finetuning while only the action expert, as well as the two prediction heads, are trained, for preserving original vision-language knowledge obtained via pre-finetuning.
\end{itemize}

Combining the two dimensions discussed above, we obtain four distinct finetuning configurations for SeqVLA:
\begin{itemize}
        \item SeqVLA-J (Joint finetuning, No Freezing).
        \item SeqVLA-JF (Joint finetuning, Freezing).
        \item SeqVLA-S (Sequential finetuning, No Freezing).
        \item SeqVLA-SF (Sequential finetuning, Freezingß).
\end{itemize}

\begin{figure}[thb]
        \centering
        \includegraphics[width=0.48\textwidth]{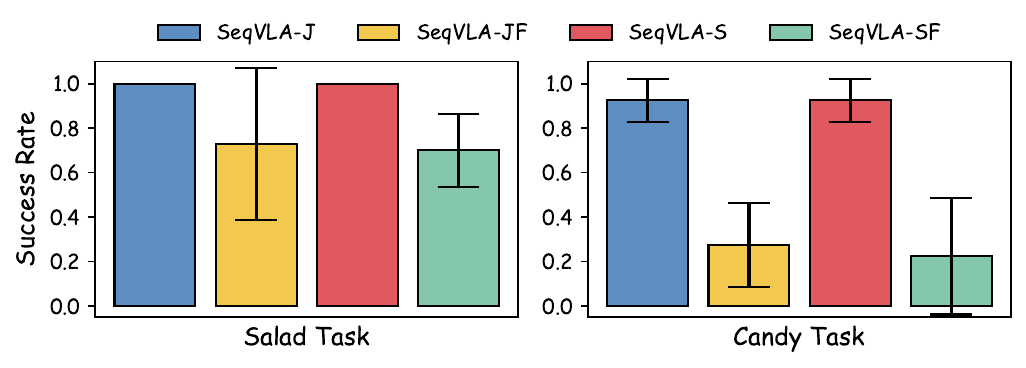}
        \caption{Finetuning strategies comparison. Average success rate for all subtasks within the two long-horizon tasks. 
        % \lz{V and J are comparatively good here but not in fig.5; action head/chuck substask}
        }
        \label{fig:training_strategy_success_rate}
\end{figure}
We instantiate four SeqVLA models corresponding to the finetuning configurations, and evaluate each model on the two long-horizon tasks. For every configuration, we compute the execution success rate over all subtasks in the salad-packing and candy-packing sequences, as summarized in \Cref{fig:training_strategy_success_rate}.  The results show that configurations with an unfrozen backbone, regardless of whether trained jointly or sequentially, consistently achieve higher success rates across subtasks. This finding highlights the importance of allowing the pre-trained backbone to adapt during finetuning, as freezing it restricts the model’s ability to transfer pre-trained representations to our domain-specific tasks.

\begin{figure}[!thb]
        % \vspace{-5mm}
        \centering
        \begin{subfigure}{0.48\textwidth}
        \centering
        \includegraphics[width=\textwidth]{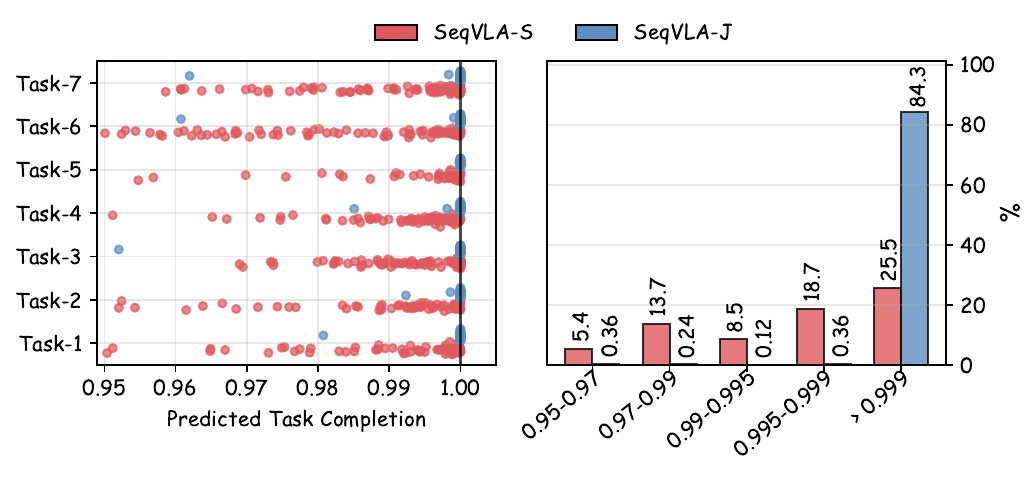}
        \vspace{-8mm}
        \caption{Salad task}
        \label{fig:task_head_salad}
        \end{subfigure}
        \begin{subfigure}{0.48\textwidth}
        \centering
        \includegraphics[width=\textwidth]{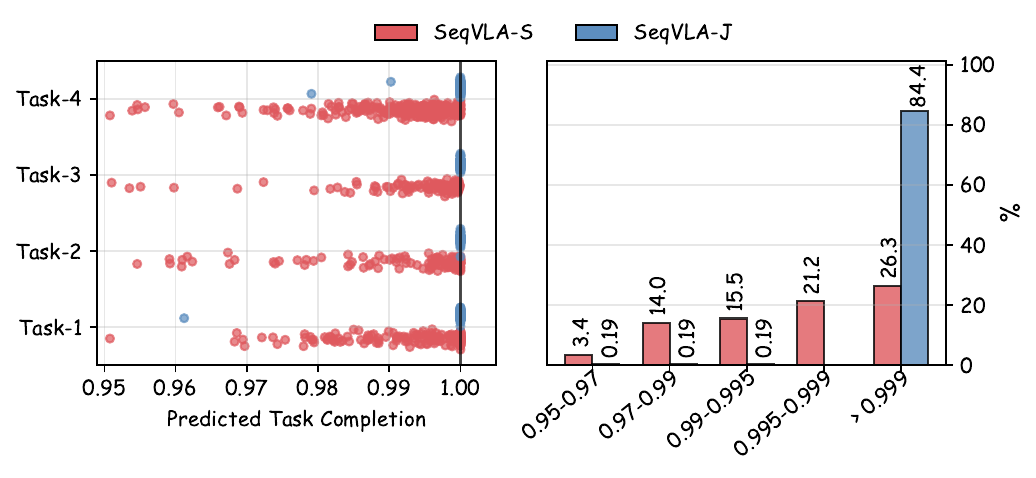}
        \vspace{-8mm}
        \caption{Candy task}
        \label{fig:task_head_candy}
        \end{subfigure}
        \caption{Task completion prediction comparison between sequential finetuning (a. SeqVLA-S) and joint finetuning (b. SeqVLA-J) strategies. Left panels show scatter plots of predicted task completion confidence for individual subtasks, where points closer to 1 indicate higher confidence in continuing execution. Right panels show success rate distributions across different confidence thresholds for (a) the Salad task with seven subtasks and (b) the Candy task with four subtasks. Overall, models trained with joint optimization (SeqVLA-J) achieve higher confidence and success rates than those trained sequentially (SeqVLA-S).} 
        \label{fig:task_head_compare}
\end{figure}

A further comparison of the completion-detection head is conducted on SeqVLA-J and SeqVLA-S, which are trained without freezing the backbone, as frozen-backbone variants already underperform substantially and are less informative for analyzing detection behavior. As shown in \Cref{fig:task_head_compare}, which displays the raw binary classification outputs, SeqVLA-J produces predictions that are both more decisive and more consistent: the output values remain much closer to 1 during task execution and drop clearly after subtask completion. In contrast, SeqVLA-S exhibits a more dispersed distribution with larger variance, often producing intermediate confidence scores that blur the boundary between execution and completion. 
\begin{table}[thb]
        \centering
        \caption{Classification Confidence Comparison}
        \vspace{-2mm}
        \label{tab:entropy_comparison}
        \begin{tabular}{lccc}
        \toprule
        \textbf{Task} & \textbf{SeqVLA-S Entropy} & \textbf{SeqVLA-J Entropy} & \textbf{KS} \\
        \midrule
        \multicolumn{4}{c}{\textit{Salad Tasks}} \\
        \midrule
        Spinach & 1.32 & \textbf{0.76} & 0.77 \\
        Cole Slaw & 1.28 & \textbf{0.70} & 0.79 \\
        Meatball & 0.79 & \textbf{0.71} & 0.57 \\
        Chicken & 1.13 & \textbf{0.81} & 0.78 \\
        Tomato & 1.47 & \textbf{0.99} & 0.69 \\
        Sauce & 1.51 & \textbf{0.57} & 0.81 \\
        Container & 1.19 & \textbf{0.48} & 0.79 \\
        \textbf{Overall} & 1.35 & \textbf{0.76} & 0.75 \\
        \midrule
        \multicolumn{4}{c}{\textit{Candy Tasks}} \\
        \midrule
        Gummies & 0.69 & \textbf{0.64} & 0.52 \\
        Kindar & 1.10 & \textbf{0.79} & 0.72 \\
        Snicker & 1.78 & \textbf{0.67} & 0.85 \\
        Lollipop & 0.83 & \textbf{0.67} & 0.80 \\
        \textbf{Overall} & 1.11 & \textbf{0.79} & 0.72 \\
        \bottomrule
        \end{tabular}

        \vspace{2mm}
        {\small \textit{Note:} All comparisons show statistical significance (p< 0.001).}
\end{table}
We quantify these differences in \Cref{tab:entropy_comparison} using the entropy of the classification outputs as a measure of confidence. Across all subtasks, SeqVLA-J consistently yields lower entropy values than SeqVLA-S, indicating that joint finetuning leads to sharper and more confident detection decisions. To further assess statistical reliability, we compute the Kolmogorov–Smirnov (KS) statistic \cite{press2007numerical} between the output distributions of execution and completion phases. SeqVLA-J achieves substantially higher KS values, with all p-values below 0.001, confirming that its classification head separates the two phases more distinctly and with a significant margin. 

Taken together, these results demonstrate that joint finetuning enables the completion-detection head to leverage shared representations with the action head, leading to more confident, statistically reliable, and practically useful subtask completion signals. In contrast, sequential finetuning tends to decouple the learning of detection from action execution, resulting in noisier predictions that may undermine reliable stage transitions in long-horizon tasks.

\begin{figure}[!thb]
        \vspace{2mm}
        \centering
        \begin{subfigure}{0.48\textwidth}
                \centering
                \includegraphics[width=\textwidth]{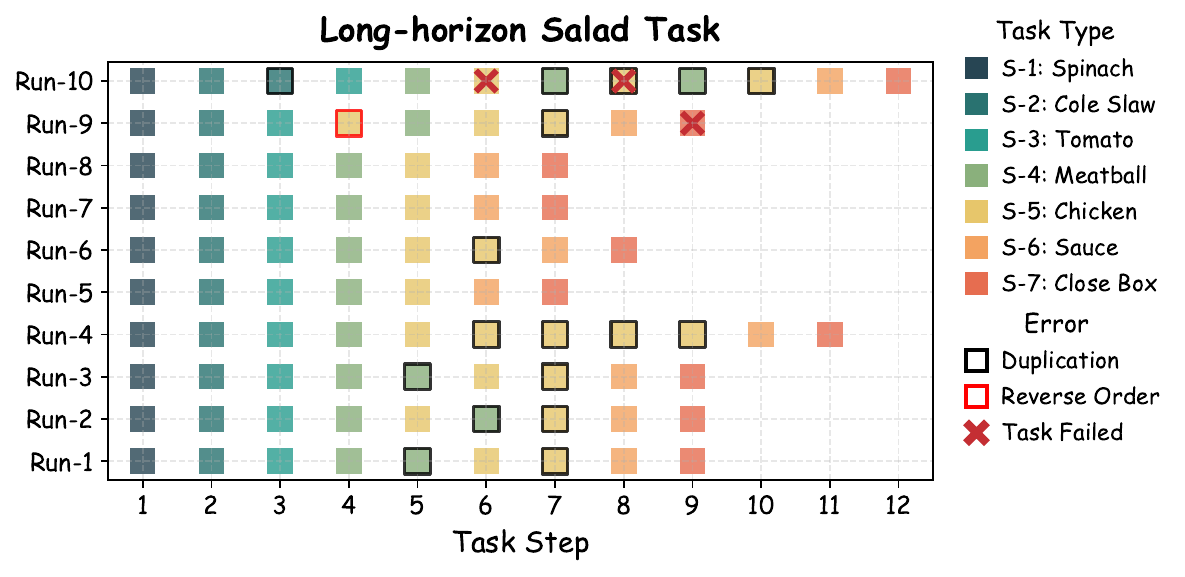}
                \label{fig:pi0_salad_task}
        \end{subfigure}
        \hfill
        \begin{subfigure}{0.48\textwidth}
                \centering
                \includegraphics[width=\textwidth]{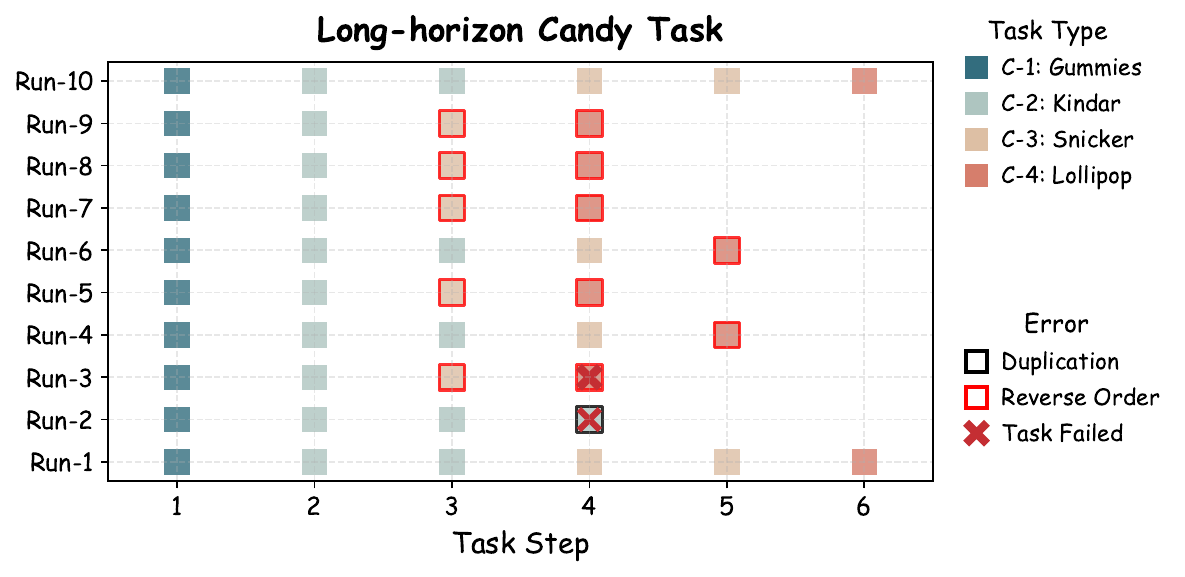}
                \label{fig:pi0_candy_task}
        \end{subfigure}
        \vspace{-10mm}
        \caption{Task execution record of baseline $\pi_0$ policy.}
        \label{fig:pi0_long_horizon_task}
\end{figure}
\begin{figure}[!th]
        \centering
        \begin{subfigure}{0.48\textwidth}
                \centering
                \includegraphics[width=\textwidth]{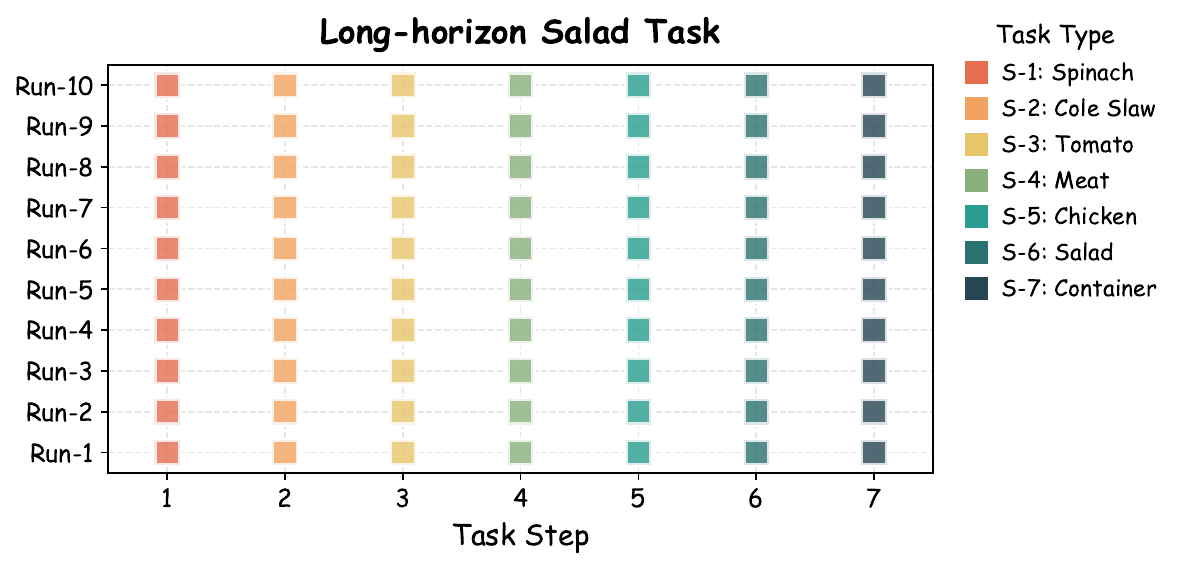}
                \label{fig:seqval_j_salad}
        \end{subfigure}
        \begin{subfigure}{0.48\textwidth}
                \centering
                \includegraphics[width=\textwidth]{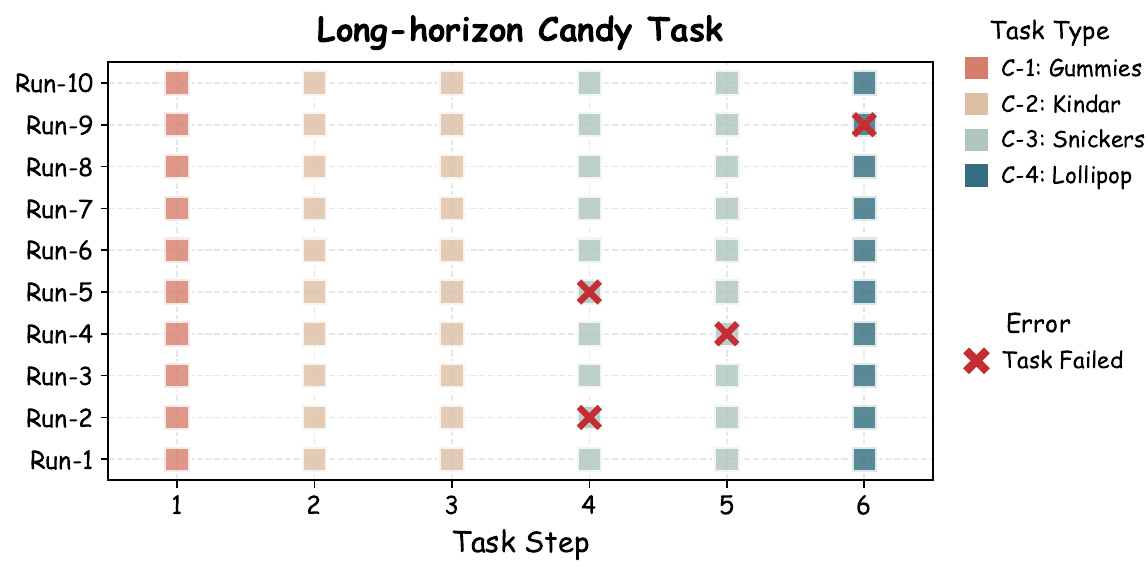}
                \label{fig:seqvla_j_candy}
        \end{subfigure}
        \vspace{-10mm}
        \caption{Task execution record of the SeqVLA-J policy.}
        \label{fig:seqvla_j_long_horizon}
\end{figure}

\begin{figure}[!thb]
        \vspace{2mm}
        \centering
        \includegraphics[width=0.48\textwidth]{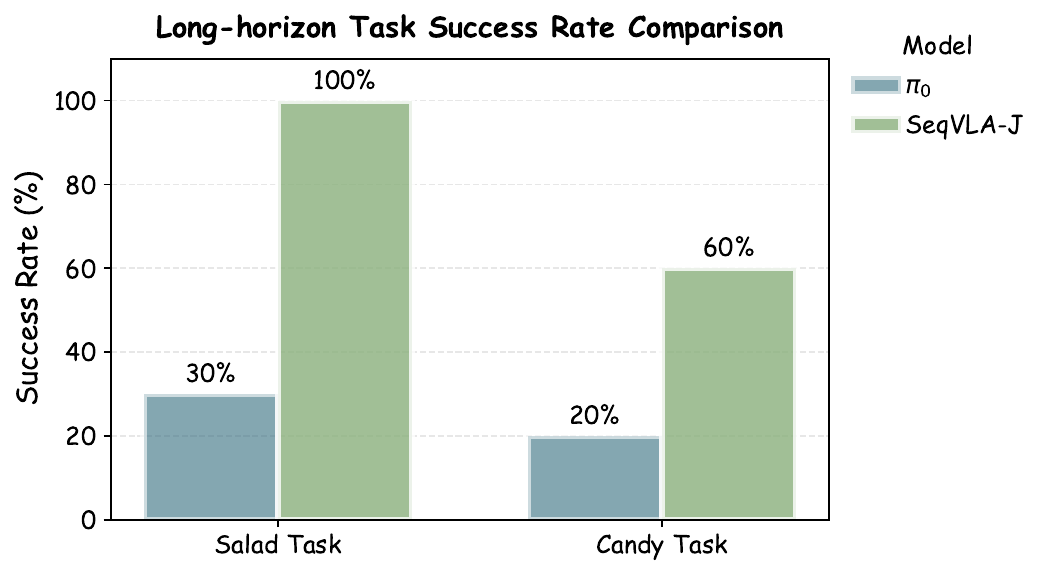}
        \caption{Success rate on long-horizon task.}
        \label{fig:pi0_vs_seqvla-j}
\end{figure}

\subsection{Long-horizon task results}
We evaluate the performance of the model trained with SeqVLA-J, i.e., the best model, on the salad and candy tasks. In addition, we also finetuned the baseline $\pi_0$ model to perform the salad and candy long-horizon tasks without subtask decomposition. The baseline model for each task is finetuned on complete sequential demonstrations (following sequences in \Cref{fig:task_sequence_diagram}), and only executes the action chunks during inference for the entire workflow, without a specific subtask monitoring scenario.

As shown in~\Cref{fig:pi0_long_horizon_task} and~\Cref{fig:seqvla_j_long_horizon}, the finetuned $\pi_0$ model achieved a comparable success rate as the SeqVLA trained on individual subtask demonstrations for both salad and candy tasks. However, the raw $\pi_0$ model struggles to maintain the proper task sequences, frequently executing subtasks out of sequence or repeating completed subtasks. In contrast, SeqVLA-J with a subtask completion detection head exhibits superior performance in executing long-horizon tasks (as shown in \Cref{fig:seqvla_j_long_horizon}), addressing the critical issue of sequential task management while maintaining overall success rates (as illustrated in \Cref{fig:pi0_vs_seqvla-j}). Although both models experience failure in some subtasks due to manipulation challenges, SeqVLA-J eliminates sequence-related failures. The dual-head architecture ensures that when failures occur, they stem from genuine manipulation difficulties rather than incorrect task ordering, making the behavior more predictable during execution.

From the success rates of the two long-horizon tasks, we observe that SeqVLA can handle long-horizon tasks that the original $\pi_0$ VLA model performs poorly on. The highly accurate performance of SeqVLA is based on $\pi_0$'s sensitivity to prompts. In particular, during finetuning, we collected 50 sets of data for each subtask and then combined them into a unified dataset. For the salad packing task, which consists of seven subtasks, the training set contained a total of 350 episodes. For the candy packing task, which consists of four distinct subtasks, the training set contained 200 episodes. We then finetuned $\pi_0$ on these aggregated datasets, enabling the model to accurately infer from the input prompt which subtask it should execute at a given stage. However, as illustrated in Figure \ref{fig:pi0_long_horizon_task},  when deployed for long-horizon tasks, the raw $\pi_0$ lacks the ability to sustain long-horizon task execution, particularly for functions with strict sequential requirements. $\pi_0$ often exhibits failures such as repeated actions or incorrect ordering. In contrast, SeqVLA mitigates these issues by incorporating a completion detection mechanism that monitors subtask status and ensures proper sequential execution, thereby enabling $\pi_0$ to accurately execute long-horizon tasks that require structured coordination across multiple subtasks.
% \begin{figure}[thb]
%         \centering
%         \begin{subfigure}{0.48\textwidth}
%                 \centering
%                 \includegraphics[width=\textwidth]{figures/long_horizon_task/seqvla_j_salad.pdf}
%                 \label{fig:seqval_j_salad_task}
%         \end{subfigure}
%         \begin{subfigure}{0.48\textwidth}
%                 \centering
%                 \includegraphics[width=\textwidth]{figures/long_horizon_task/seqvla_j_candy.pdf}
%                 \label{fig:seqvla_j_candy_task}
%         \end{subfigure}
%         \vspace{-10mm}
%         \caption{Task excution record of SeqVLA-J models.}
%         \label{fig:seqvla_j_long_horizon_task}
% \end{figure}

\section{CONCLUSIONS and FUTURE WORK}
\subsection{Conclusion}
This work addresses the challenge of long-horizon sequential manipulation, where existing VLA models, such as $\pi_0$, lack explicit signals for subtask completion. We proposed SeqVLA, which augments $\pi_0$ with a lightweight completion detection head to generate actions and subtask transitions jointly. By evaluating four finetuning strategies that balance dual-head optimization and pre-trained knowledge preservation, we showed that SeqVLA achieves higher success rates than the baseline across salad- and candy-packing benchmarks. Notably, joint finetuning with an unfrozen backbone yielded the most confident and reliable completion predictions. These results highlight that coupling action generation with explicit subtask-aware detection is a promising direction for scaling VLA models to complex sequential manipulation tasks.

\subsection{Future Work}
Several promising directions emerge from this work that warrant further investigation. A natural extension is to move beyond linear task sequences toward hierarchical structures with conditional branching and parallel execution. Instead of a binary detection head, future models could employ multi-level state detection to track progress across subtasks and dependencies, enabling more sophisticated workflows such as multi-dish preparation or dual-arm coordination, where subtasks may run concurrently or adapt their order to environmental conditions.

A further extension is to explore human–robot collaboration. In many realistic scenarios, robots will work alongside human operators on long-horizon tasks. Extending our framework to such settings would require mechanisms for effective communication, shared understanding of subtask states, and smooth handover of responsibilities between human and robotic execution.

Finally, real-world deployment beyond laboratory conditions presents its own challenges, including safety constraints, fault tolerance, and integration into existing workflows. Robust error recovery mechanisms, particularly for unexpected subtask failures, will be essential to ensure reliable operation in environments such as restaurants, homes, or industrial facilities. Addressing these issues will be critical to bridging the gap between controlled experiments and practical, large-scale deployment.


% \section*{ACKNOWLEDGMENT}
%%%%%%%%%%%%%%%%%%%%%%%%%%%%%%%%%%%%%%%%%%%%%%%%%%%%%%%%%%%%%%%%%%%%%%%%%%%%%%%%

\bibliography{refs}

\end{document}